\documentclass[10pt,conference]{IEEEtran}

\usepackage[T1]{fontenc}\usepackage[utf8]{inputenc}\usepackage{newtxtext,newtxmath}
\usepackage{graphicx,booktabs,tabularx,array,amsmath,bm,algorithm,algpseudocode,microtype,url}
\usepackage[hidelinks]{hyperref}
\usepackage{fancyhdr}
\fancypagestyle{plain}{
  \fancyhf{}
  \fancyfoot[L]{\footnotesize Mentomy AI}
  \fancyfoot[C]{\thepage}

}

\graphicspath{{figures/}}\emergencystretch=1em

\newcommand{\Qset}{\mathcal{Q}}\newcommand{\state}{\bm{s}}\newcommand{\weights}{\bm{w}}
\newcommand{\Ev}{\mathcal{E}_{v}}\newcommand{\Ec}{\mathcal{E}_{c}}

\title{An Auditable Symbolic--RAG--Generative AI Architecture for Goal-Oriented Conversation Orchestration}

\author{\IEEEauthorblockN{Ramón González and Antonio Díaz}
\IEEEauthorblockA{Department of AI Research, Mentomy AI\\
Almería, Spain\\
ramon@mentomy.com}}

\begin{document}\maketitle
\thispagestyle{plain}
\pagestyle{plain}

\begin{abstract}
Goal-oriented conversational systems must answer factual questions, understand visitor-provided
information, and advance business objectives without becoming rigid questionnaires. This paper
proposes a Symbolic--RAG--Generative architecture centered on the Goal-oriented
Retrieval-Augmented Conversation Engine (GRACE). An instruction-constrained Business Goal Compiler transforms business
intent into an immutable objective set, normalized priority vector, canonical questions, and
initial state vector. At runtime, GRACE receives the complete conversation history, latest
visitor message, current state, and grounded answer generated by a separate RAG component. It
updates completion only from visitor-authored evidence and selects one contextually
modulated follow-up. The core policy maximizes expected business progress subject to a minimum
visitor-utility constraint. We formalize the state, monotonic transitions, source separation,
question modulation, and constrained policy; present the reference architecture; and define
an evaluation comprising 24 English real-estate and 10 Spanish professional-cleaning conversations, totaling 119
protocol-defined visitor turns. Across both domains, GRACE achieves 84.9\% exact state-transition
accuracy, 91.6\% evidence precision, 89.6\% evidence recall, 100\% monotonicity, and 94.1\%
terminal-state accuracy.
The evaluation establishes compelling symbolic-state performance across standard, multi-goal, RAG-detour,
validation, refusal, and robustness scenarios.
\end{abstract}

\section{Introduction}
Conversational interfaces increasingly combine enterprise knowledge with business workflows.
A visitor may ask whether properties are available near a city, state a budget and bedroom
requirement in one message, and later ask whether international schools are nearby.
The system must answer from governed knowledge, remember what the visitor already supplied,
and decide whether one follow-up is appropriate. This is controlled conversation orchestration,
not merely response generation.

Existing approaches occupy different points in the design space. Classical chatbots and
slot-filling systems execute predefined questions efficiently but depend on engineered states,
intents, and dialogue policies \cite{young2013pomdp,williams2016dstc}. RAG assistants answer
open questions from retrieved knowledge \cite{lewis2020rag}, but retrieval alone does not
define which business information remains unknown. General LLM agents reason and invoke tools
across broad action spaces \cite{yao2023react,schick2023toolformer}, adding flexibility but also
behavioral variance and iterative latency. Industrial task-oriented dialogue work likewise
highlights the tension among adaptability, conversational quality, and business constraints
\cite{poddar2022industry}. GRACE targets a bounded middle ground: generative understanding and
natural phrasing within an explicit, auditable goal-state envelope.

The contributions are: (1) two instruction-constrained roles separating goal compilation from
runtime orchestration; (2) a persistent symbolic completion vector with immutable semantics and
monotonic transitions; (3) explicit separation between visitor evidence and company-grounded
answers; (4) grounded-answer-conditioned modulation of canonical questions; (5) an auditable
turn record linking each accepted transition to its visitor-authored witness; and (6) an
$\epsilon$-constrained policy that maximizes expected business progress only among actions
meeting a minimum visitor-utility requirement.

Table~\ref{tab:position} positions GRACE architecturally rather than as an experimental baseline.
It contrasts the control, flexibility, state representation, and expected latency of four
conversation paradigms.
\begin{table*}[!ht]\caption{Conceptual comparison of conversational-system architectures}\label{tab:position}\centering\scriptsize
\begin{tabular}{p{.12\textwidth}p{.17\textwidth}p{.17\textwidth}p{.17\textwidth}p{.18\textwidth}}\toprule
Dimension & Closed-question chatbot & RAG assistant & General AI agent & GRACE \\
\midrule
Primary function & Execute a fixed flow & Answer from retrieved knowledge & Plan and execute open-ended actions & Answer through RAG while progressing approved goals \\
Understanding & Pattern, intent, or slot dependent & Strong open-language question understanding & Strong reasoning and tool interpretation & Open-language interpretation constrained by explicit evidence rules \\
Flexibility & Low & High for answering; limited goal tracking & Very high but broad & High inside a bounded objective space \\
State & Branch position or slots & History or latent context & Memory and plan state & Explicit $\state_t$, evidence witnesses, and immutable $\Qset$ \\
Latency & Low & Retrieval plus synthesis & Often iterative and high & Retrieval/synthesis plus one bounded orchestration step \\
Auditability & High for scripted branches & Evidence can be cited; goal progress is implicit & Action reasoning may vary & Goal, state, evidence source, and transitions are explicit \\
Representative work & \cite{young2013pomdp,williams2016dstc} & \cite{lewis2020rag,shuster2021retrieval} & \cite{yao2023react,schick2023toolformer} & This work \\
\bottomrule\end{tabular}\end{table*}

\section{Literature Review}
\subsection{Dialogue State Tracking}
Dialogue State Tracking (DST) summarizes dialogue history into a state used by a downstream
policy. The DST Challenge
established shared tasks and metrics \cite{williams2016dstc}; surveys document the transition
from discriminative trackers to generative text-to-text models \cite{jacqmin2022dst}.
MultiWOZ enabled multi-domain evaluation \cite{budzianowski2018multiwoz}, TRADE studied
transfer \cite{wu2019trade}, and schema-driven prompting showed that natural-language schema
descriptions can guide general language models \cite{lee2021schema}. LLM-driven and context-
aware prompting methods further address zero-shot and cross-domain tracking
\cite{feng2023llmdst,dong2024capid}. The Schema-Guided Dialogue benchmark tests transfer across
changing service schemas \cite{rastogi2020sgd}, while dataflow representations retain explicit,
revisable conversational structure beyond flat slot sets \cite{andreas2020dataflow}.
Multilingual resources such as PRESTO further show that realistic task-oriented conversations
include language switching, revisions, and structured context that simplified benchmarks often omit
\cite{goel2023presto}. More recently, Carranza and Rojas represented dialogue state as an
LLM-generated natural-language summary to improve interpretability and robustness
\cite{carranza2025nldst}, while ReacTOD combined agentic inference with deterministic symbolic
validation for bounded, zero-shot DST \cite{lin2026reactod}.

GRACE retains DST's explicit-state principle but changes the state semantics. Conventional DST
tracks domain--slot--value constraints needed to satisfy the user's task. GRACE tracks whether
the company has obtained sufficient visitor-authored evidence for each approved objective.
It adds immutable priorities, monotonic completion, refusal handling, and an evidence witness
for every transition.

\subsection{End-to-End and Retrieval-Grounded Dialogue}
End-to-end task-oriented dialogue combines tracking, policy, database interaction, and response
generation \cite{qin2023e2e}. Prompting, retrieval, and fine-tuning offer different trade-offs
under distribution shift \cite{raposo2023prompting}. RAG combines parametric generation and
non-parametric memory \cite{lewis2020rag}; dense retrieval supports large knowledge bases
\cite{karpukhin2020dpr}. Retrieval-generation dialogue architectures improve knowledge access
and response diversity \cite{nekvinda2022aargh,shi2023dualfeedback}. Retrieval augmentation can
reduce conversational hallucination \cite{shuster2021retrieval}, while attribution benchmarks
show that factual overlap alone is insufficient: a response must be supported by its identified
source \cite{dziri2022begin}. Recent work has expanded RAG along different axes. DH-RAG
incorporates dynamically maintained dialogue history into query reconstruction for multi-turn
retrieval \cite{zhang2025dhrag}, and a recent survey organizes advances around retrieval,
generation, robustness, and efficiency \cite{sharma2025ragsurvey}. GRACE instead keeps the
demonstrated retrieval interface deliberately turn-local: company passages are retrieved from the
current visitor message, while conversation history is reserved for orchestration and
visitor-evidence validation.

GRACE contributes a provenance boundary absent from ordinary answer-grounding objectives:
retrieved company facts may condition a follow-up, but cannot be treated as facts learned
about the visitor. Factual synthesis and business-state progression remain separate.

\subsection{Policy, Agents, and Symbolic Constraints}
Dialogue policy has been modeled using MDPs, POMDPs, and hierarchical reinforcement learning
\cite{young2013pomdp,levin2000dialog,peng2017hierarchical}. General LLM agents broaden the
action space through interleaved reasoning and tool use \cite{yao2023react,schick2023toolformer}.
Within dialogue systems, hybrid agents have used LLMs to contextualize pipeline decisions,
handle out-of-scope input, and formulate clarification questions while retaining established
controls \cite{foosherian2023enhancing}. Self-talk has been used to generate and filter training
dialogues for specialized task-oriented agents \cite{ulmer2024selftalk}, while induced dialogue
flows constrain LLM chatbots to domain-relevant trajectories \cite{agrawal2024dialogflow}.
These cases span tool-using, pipeline-augmented, trained, and flow-constrained agents. GRACE is
distinct in permitting one approved question or NONE, separating company-grounded facts from
visitor evidence, and applying an explicit visitor-utility floor before optimizing business
progress. NeuSymMS provides a complementary neuro-symbolic design in which LLM-extracted facts are
governed by explicit lifecycle rules in persistent agent memory \cite{sultan2026neusymms}. GRACE
differs in scope: its symbolic record represents completion of an approved, conversation-level
objective set, and its source constraint determines which visitor evidence may change that record.

Across these strands, prior work primarily develops state estimation, grounded response
generation, dialogue policy, or structural validity. GRACE's contribution is their constrained
composition: compiled business objectives define the symbolic state, RAG supplies company facts,
and the orchestrator advances an objective only from visitor-authored evidence. This is an
architectural distinction, not a claim of empirical superiority over the cited methods.

\section{Problem Formulation}\label{sec:problem}
\subsection{Objective Set, State, and Conversation History}
Before runtime orchestration begins, the instruction-constrained Business Goal Compiler converts
a manager's business intention into a stable symbolic specification. This subsection formalizes
the Compiler's principal outputs and the conversation information consumed by GRACE.
The Compiler produces
\begin{equation}\Qset=\{q_i\}_{i=1}^{n},\quad \weights\in\mathbb{R}_{+}^{n},\quad \sum_{i=1}^{n}w_i=1,\label{eq:goal}\end{equation}
where $q_i$ includes a stable identifier, canonical question, completion criterion, and weight.
The CurrentState is
\begin{equation}\state_t=[s_{t,1},\ldots,s_{t,n}]^{\top}\in\{0,1\}^{n}.\label{eq:state}\end{equation}

The history $H_t$ is the complete ordered conversation before the current decision. It includes
all visitor messages and every AI-visible response. An AI response consists of the grounded
RAG answer and the question or NONE returned by GRACE. Thus,
\begin{equation}H_t=\big[(u_1,y_1),\ldots,(u_{t-1},y_{t-1})\big],\quad y_k=(r_k,m_k),\label{eq:history}\end{equation}
where $u_k$ is the visitor turn, $r_k$ the RAG answer, and $m_k$ the orchestrator output:
one question or NONE. This definition prevents ``history'' from being misread as visitor text alone.

For any binary state $\state$, the feasible action set is
\begin{equation}\mathcal{A}(\state)=\{\mathrm{NONE}\}\cup\{\mathrm{ASK}(i,m_{t,i}):s_i=0\}.\label{eq:action}\end{equation}
The canonical question defines what must be learned. Its wording is modulated by the grounded
answer and conversation context:
\begin{equation}\begin{aligned}m_{t,i}&=M_{\mathrm{LLM}}(q_i,u_t,H_t,r_t),\\
\operatorname{meaning}(m_{t,i})&\equiv\operatorname{meaning}(q_i).\end{aligned}\label{eq:modulation}\end{equation}
For example, after a RAG answer gives a price of 325.000 EUR, ``What budget are you considering?''
may become ``Is 325.000 EUR around the budget you had in mind?'' Figure~\ref{fig:runtime}
later shows where $r_t$ enters this modulation process.

\subsection{Retrieval Grounding and Provenance}
The RAG component produces a company-grounded answer from the current visitor message and passages
retrieved from the company's frozen corpus. It does not use the dialogue history or the symbolic
state. Formally, the synthesizer is
\begin{equation}
r_t=G_{\mathrm{RAG}}(u_t,\mathcal D_t),
\label{eq:ragsynthesis}
\end{equation}
where $u_t$ is the current visitor message and $\mathcal D_t$ is the retrieved evidence set.
Let $\mathcal C$ denote the company corpus, $E$ the multilingual embedding model, and $K$ the
configured retrieval depth. The evidence supplied to Equation~\eqref{eq:ragsynthesis} is
\begin{equation}
\mathcal D_t=\operatorname{TopK}_{d\in\mathcal C}
\operatorname{sim}\!\left(E(u_t),E(d)\right).
\label{eq:retrieval}
\end{equation}
Here, $\operatorname{sim}:\mathbb R^p\times\mathbb R^p\rightarrow\mathbb R$ denotes the configured
vector-similarity score used to rank two $p$-dimensional embeddings; larger values indicate greater
semantic similarity.
A grounded company claim must be supported by at least one retrieved passage:
\begin{equation}
c\in\operatorname{Claims}(r_t)\Rightarrow
\exists d\in\mathcal D_t:\ d\models c.
\label{eq:grounding}
\end{equation}
Equations~\eqref{eq:retrieval}--\eqref{eq:grounding} specify the interface required by GRACE;
they do not propose a new retrieval algorithm. Most importantly, $\mathcal D_t$ and $r_t$ belong
to company evidence $\Ec$, never visitor evidence $\Ev$. They may determine the factual answer and
modulate a follow-up, but cannot complete a visitor objective.

\subsection{Source-Constrained Monotonic Transition}
Let $\Ev(H_t,u_t)$ contain visitor-authored propositions and $\Ec(H_t,r_t)$ contain AI or
company-grounded propositions. Define $\bm z_t=[z_{t,1},\ldots,z_{t,n}]^\top$ as the vector of
newly accepted completion events. A positive indicator requires an unfinished coordinate and
visitor evidence satisfying its published criterion:
\begin{equation}z_{t,i}=1\Rightarrow s_{t,i}=0\ \land\ \exists e\in\Ev(H_t,u_t):e\models\operatorname{criterion}(q_i).\label{eq:source}\end{equation}
The accepted transition is
\begin{equation}s_{t+1,i}=\max(s_{t,i},z_{t,i}),\quad z_{t,i}\in\{0,1\},\label{eq:transition}\end{equation}
and therefore
\begin{equation}s_{t+1,i}\ge s_{t,i},\quad \|\state_{t+1}\|_1\ge\|\state_t\|_1.\label{eq:monotonic}\end{equation}
Equation~\eqref{eq:monotonic} expresses monotonicity: once completion is valid, the
flag cannot return from 1 to 0. The count of completed objectives stays equal or increases.
The present binary model addresses evidence accumulation, not later withdrawal. Future work will
consider non-binary states that can represent changed, revoked, or uncertain preferences over a
conversation.

\subsection{Visitor-Constrained Business Policy}
The reference policy does not claim a learned elicitation probability. Because $\mathrm{ASK}(i,m)$
targets one approved objective, its business score is the transparent deterministic priority
\begin{equation}U_B(a)=\begin{cases}w_i,&a=\mathrm{ASK}(i,m),\ s_{t+1,i}=0,\\0,&a=\mathrm{NONE}.
\end{cases}\label{eq:business}\end{equation}
Consequently, among visitor-admissible questions, GRACE asks the highest-weight unsatisfied goal;
equal scores are resolved by the Compiler's approved order. This is intentionally simpler and
directly reproducible: business utility is the published priority of the unfinished objective
targeted by the action. More elaborate learned scoring is outside the scope of this study.

Visitor utility uses a common normalized rubric. Each component belongs to $\{0,0.5,1\}$ and
is emitted with a short rationale: relevance $R$, interaction burden $B$, repetition $D$, and
privacy sensitivity $P$. The reference configuration is
\begin{equation}U_V(a)=0.40R(a)-0.20B(a)-0.20D(a)-0.20P(a),\quad \epsilon=0.\label{eq:visitor}\end{equation}
The levels mean absent, moderate, or strong evidence for the corresponding property. The
coefficients are an explicit policy choice, not learned quantities; an organization may change
them only through a versioned, approved configuration. The reference value $\epsilon=0$ admits
only actions whose weighted relevance is at least as large as their combined burden, repetition,
and privacy penalties; it is a transparent neutral-utility floor, not an empirically calibrated
preference threshold. Refusal or a direct request to stop sets
the relevant ASK action below the threshold independently of its business priority.
The no-question action is assigned $U_V(\mathrm{NONE})=0$, so it remains feasible at the reference
threshold and the constrained decision always has at least one admissible action.

\begin{figure}[!ht]\centering\includegraphics[width=\columnwidth]{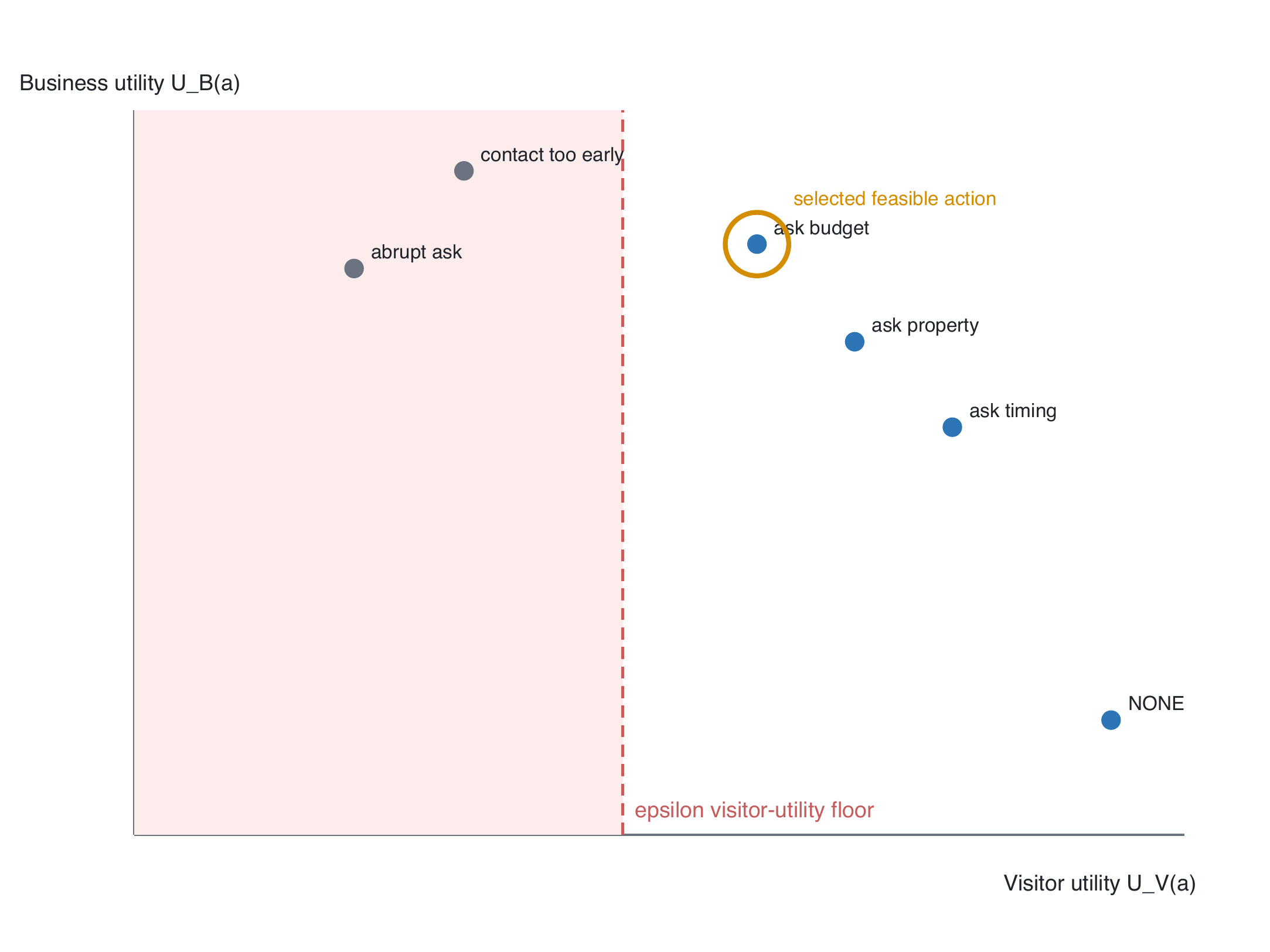}
\caption{Schematic $\epsilon$-constrained policy of Equation~\eqref{eq:core}; $\epsilon=0$ in the reference configuration and point locations are illustrative.}\label{fig:epsilon}\end{figure}

The core decision is
\begin{equation}\begin{aligned}a_t^*=\arg\max_{a\in\mathcal A(\state_{t+1})}\;&U_B(a)\\
\mathrm{subject\ to}\quad &U_V(a)\ge\epsilon.\end{aligned}\label{eq:core}\end{equation}
A Pareto trade-off exists because improving information acquisition can reduce visitor
experience. Equation~\eqref{eq:core} resolves the trade-off lexically: exclude every action
below the visitor-utility floor $\epsilon$, then choose the feasible action with maximum
business utility. Figure~\ref{fig:epsilon} visualizes this rule. Points left of the threshold
are infeasible even when their business utility is high; the highlighted point is selected
because it has the greatest $U_B$ among feasible actions.

\section{Reference Architecture and Operational Specification}\label{sec:architecture}
\subsection{Business Goal Compiler}
\begin{figure}[!ht]\centering\includegraphics[width=\columnwidth]{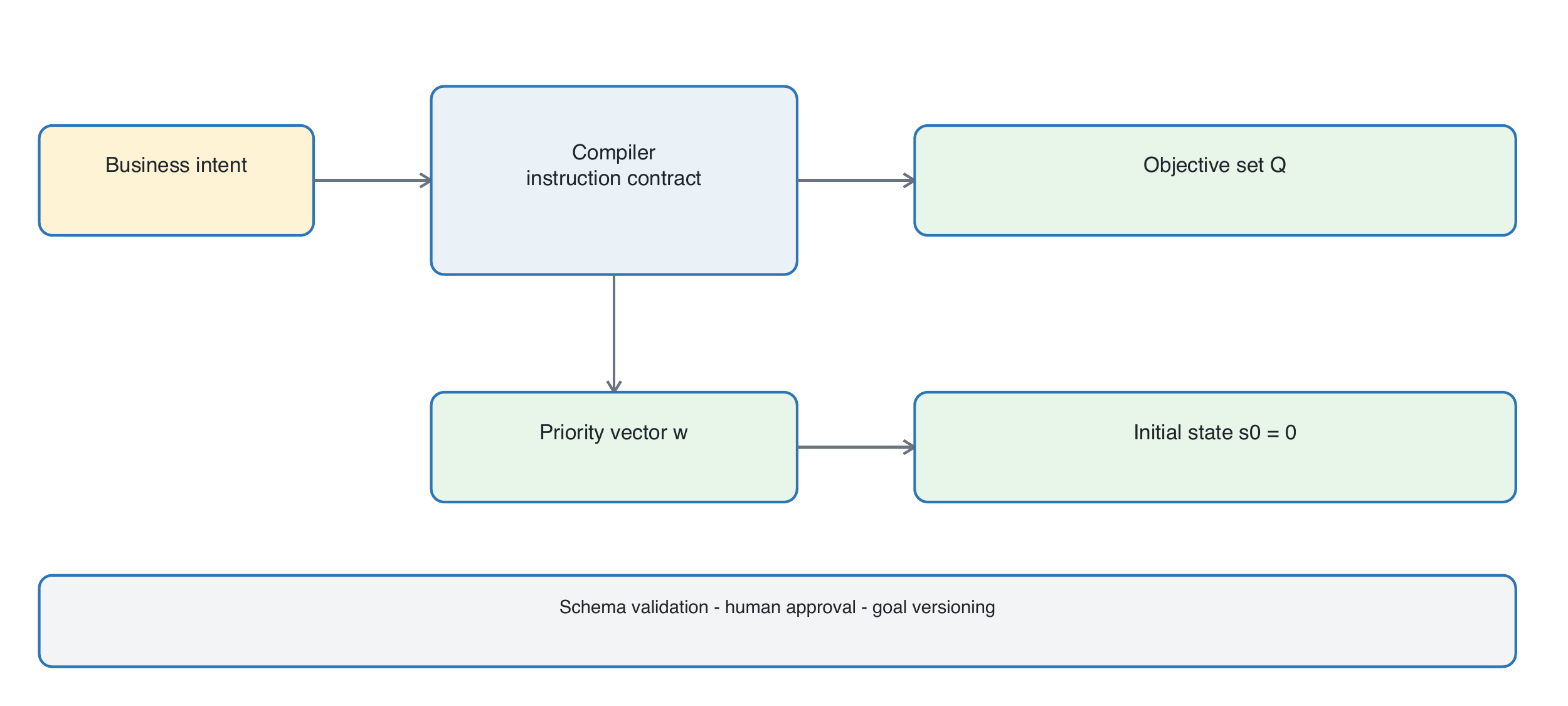}
\caption{Offline Business Goal Compiler and validated symbolic outputs.}\label{fig:compiler}\end{figure}

The architecture in Figure~\ref{fig:compiler} begins offline. An instruction-constrained
Compiler transforms the manager's business intention into $\Qset$, $\weights$, evidence
criteria, and $\state_0$. The diagram separates semantic compilation from validation and human
approval; this keeps goal semantics stable during a conversation.

Algorithm~\ref{alg:compiler} makes the compilation sequence explicit. It operationalizes
Equation~\eqref{eq:goal}: an instruction-constrained Compiler model generates a structured candidate,
after which deterministic validation and manager approval establish the published goal version.
The evaluated model and decoding configuration are reported in Section~\ref{sec:design}.
\begin{algorithm}[!ht]\caption{Instruction-Constrained Goal Compilation}\label{alg:compiler}\begin{algorithmic}[1]
\Require Manager intention $\mathcal I$, governance specification $\Gamma$
\State Invoke Compiler with $\mathcal I$, $\Gamma$, output schema, and examples
\State Receive candidate $(\Qset,\weights,\state_0)$ and criteria
\State Validate uniqueness, normalization, and permitted data
\State Obtain human approval; version and publish
\end{algorithmic}\end{algorithm}

The Compiler output specification requires a stable identifier, goal name, canonical question,
completion criterion, priority weight, and approved order for every goal.
Its structured output is rejected if identifiers are duplicated, fields are absent, weights do
not sum to one, or a criterion requests prohibited data. These checks are deterministic; human
approval establishes the published goal version.

\subsection{GRACE Runtime}
\begin{figure}[!ht]\centering\includegraphics[width=\columnwidth]{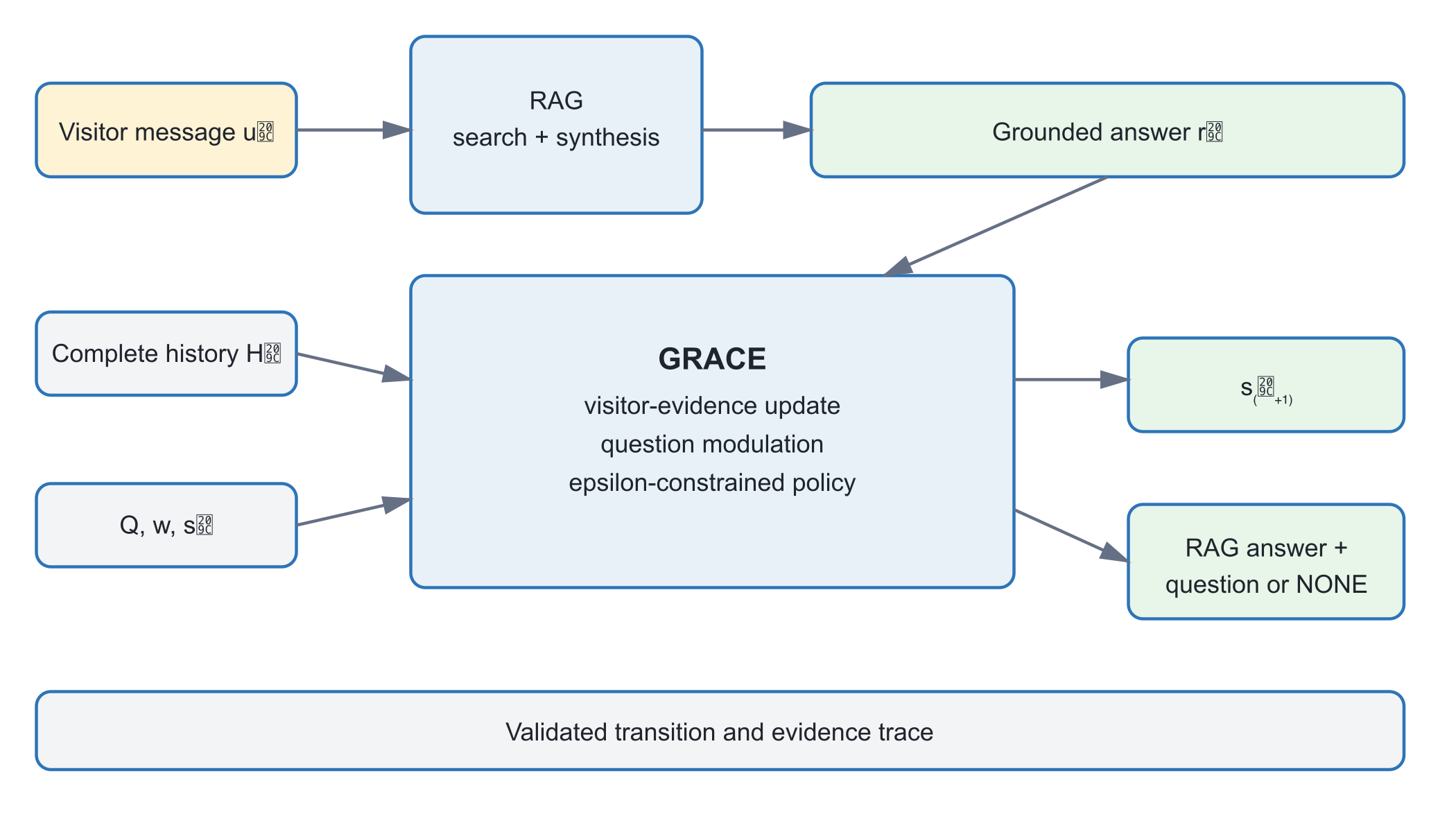}
\caption{Online GRACE flow and the inputs included in the conversation history.}\label{fig:runtime}\end{figure}

Figure~\ref{fig:runtime} shows the online data flow. The RAG component first synthesizes $r_t$.
GRACE then receives $r_t$ together with $H_t$, $u_t$, $\Qset$, $\weights$, and $\state_t$.
The grounded answer influences question modulation but not completion provenance. GRACE emits
the new validated state $\state_{t+1}$ and one question or NONE; the visible AI response is the
composition of the RAG answer and the GRACE output. Here, $\state_t$ is the validated state entering
turn $t$, whereas $\state_{t+1}$ is the new state after processing and validating the evidence in
$u_t$ against the visitor-authored conversation record.
In the reference deployment, embedding, semantic retrieval, and grounded synthesis execute on the
RAG service connected to Pinecone. The orchestration model executes independently and receives the
completed RAG result rather than querying the vector database itself. This separation permits load
balancing and prevents retrieval latency or corpus access from being hidden inside symbolic state
updates.

Algorithm~\ref{alg:runtime} links the runtime architecture to Equations~\eqref{eq:source}--
\eqref{eq:core}. It is the turn-level procedure evaluated in Section~\ref{sec:results}.
\begin{algorithm}[!ht]\caption{GRACE Turn}\label{alg:runtime}\begin{algorithmic}[1]
\Require $\Qset,\weights,\state_t,H_t,u_t,\mathcal C,E,K,G_{\mathrm{RAG}},U_V,\epsilon$
\State Retrieve $\mathcal D_t$ from $\mathcal C$ using Equation~\eqref{eq:retrieval}
\State Synthesize grounded answer $r_t$ using Equation~\eqref{eq:ragsynthesis}
\State If $\mathcal D_t$ provides no usable support, set $r_t$ to an explicit unavailable-information response
\State Identify visitor-only evidence and derive $\bm z_t$
\State Validate the evidence and form the new state $\state_{t+1}=\state_t\lor\bm z_t$
\State Generate faithful question modulations $m_{t,i}$
\State Apply Equation~\eqref{eq:core}; select one question or NONE
\State Persist the new state and transition record
\State Return $r_t$; append the selected question only if $a_t^*$ is an ASK action
\end{algorithmic}\end{algorithm}

The runtime output consists of the new state, the selected action, and the response text. Validation
checks schema conformance, immutable goal identifiers and weights, legal monotonic transitions,
visitor support for newly completed coordinates, and contact syntax when applicable. If validation
fails, the accepted state remains unchanged.

\section{Evaluation Methodology}\label{sec:design}
\subsection{Experimental Setup}
We evaluated two complete demonstrations: an English-language real-estate lead-qualification demo
and a Spanish-language professional-cleaning demo. Together they contain 34 scripted conversations
and 119 protocol-defined visitor turns. The Business Goal Compiler, GRACE, and the RAG answer
synthesizer use Mistral Small 4 (\texttt{mistral-small-2603}), released on March 16, 2026
\cite{mistral2026small4}, with temperature $0.2$ and a maximum output length of 4,096 tokens.
Retrieval uses a Pinecone vector database for semantic search \cite{pinecone2026semantic} and the
\texttt{intfloat/multilingual-e5-large} embedding model \cite{wang2024multilingual}.

Both demonstrations ran in Mentomy, a private, sovereign AI platform designed to keep company
knowledge and conversational workflows within a governed deployment. The experiments used
Mentomy's Widget functionality, through which a company embeds an AI chat interface in its website.
The Widget combines the company's RAG-backed knowledge service with GRACE's goal-oriented runtime,
so visitors can ask company questions naturally while the approved symbolic objectives progress in
the same interaction.

The model version, decoding settings, Compiler and GRACE instructions, goal specification, canonical
order, RAG corpus, embedding model, vector index, retrieval parameters, and validator version were
frozen within each demo. The two Compiler outputs share
\begin{equation}
\weights=(0.35,0.25,0.20,0.20),\qquad \state_0=[0,0,0,0],
\label{eq:evalconfig}
\end{equation}

\subsection{Performance Metrics}
The visitor turn is the primary unit of analysis. Let $\mathcal T$ denote the set of evaluated
turns, $\widehat{\state}_{t+1}$ the state vector recorded by the system after turn $t$, and
$\state^{\star}_{t+1}$ the reference state derived exclusively from visitor-authored evidence.
State-transition accuracy is
\begin{equation}
\mathrm{STA}=\frac{1}{|\mathcal T|}\sum_{t\in\mathcal T}
\mathbb 1[\widehat{\state}_{t+1}=\state^{\star}_{t+1}].
\label{eq:sta}
\end{equation}
STA is meaningful because it requires all coordinates to be correct at the exact turn where the
visitor evidence appears; a correct terminal vector cannot conceal a delayed or misplaced update.

Let $\mathcal L$ denote the evaluated conversations and let
$\widehat{\state}^{(\ell)}_{\mathrm{end}}$ and $\state^{\star(\ell)}_{\mathrm{end}}$ be their recorded
and reference terminal states. Terminal-state accuracy is
\begin{equation}
\mathrm{TSA}=\frac{1}{|\mathcal L|}\sum_{\ell\in\mathcal L}
\mathbb 1[\widehat{\state}^{(\ell)}_{\mathrm{end}}=\state^{\star(\ell)}_{\mathrm{end}}].
\label{eq:terminal}
\end{equation}
TSA measures whether each conversation ultimately preserves the complete evidence-derived outcome,
whereas STA measures whether it reaches that outcome through the correct turn-level transitions.

Monotonicity compliance is
\begin{equation}
\mathrm{MC}=\frac{1}{|\mathcal T|}\sum_{t\in\mathcal T}
\mathbb 1[\widehat{\state}_{t+1}\succeq\widehat{\state}_{t}].
\label{eq:metricmonotonic}
\end{equation}
MC tests the binary-state invariant in Equation~\eqref{eq:monotonic}. It is reported separately
because an unsupported transition can still be monotonic.

For evidence-related metrics, let $\mathcal Z$ be the set of coordinate-completion events recorded
by GRACE and $\mathcal Z^{\star}$ the reference set supported by visitor evidence. A
coordinate-completion event is a conversation--turn--objective tuple $(\ell,t,i)$ at which
coordinate $i$ changes from incomplete to complete; one visitor turn can therefore contain several
events. Evidence precision is
\begin{equation}
\mathrm{EP}=\frac{|\mathcal Z\cap\mathcal Z^{\star}|}{|\mathcal Z|}.
\label{eq:precision}
\end{equation}
EP measures how often a recorded completion is justified at the correct turn and therefore
penalizes unsupported or temporally misattributed transitions.

Evidence recall is
\begin{equation}
\mathrm{ER}=\frac{|\mathcal Z\cap\mathcal Z^{\star}|}{|\mathcal Z^{\star}|}.
\label{eq:recall}
\end{equation}
ER measures how much of the completion evidence actually supplied by visitors is captured at the
correct turn by the state update.

\section{Demonstration Configurations}\label{sec:cases}
The same frozen architecture was instantiated in two domains and languages. Each company was
simulated for experimental purposes, and each supplied a plain-language business intention and a
company-authored FAQ collection from which its vector database was generated. Rather than imitate
routine conversations, the scripts concentrate difficult behaviors into predefined scenario
families. This increases diagnostic density and avoids the privacy concerns of collecting real
prospect conversations.

\subsection{English Real-Estate Demonstration}\label{sec:case}
The manager supplied the following plain-language intention: ``We want to know the client's budget,
area of interest, housing requirements, and contact details in order to arrange a joint viewing.''
The Compiler transformed it into the four goals in Table~\ref{tab:compileroutput}; their coordinate
order is budget, location, requirements, and contact.

\begin{table*}[!t]\caption{Real-Estate Goals Produced by the LLM Compiler}\label{tab:compileroutput}\centering\small
\begin{tabular}{clp{.47\textwidth}cc}\toprule
Order & Objective & Canonical question produced by the Compiler & Weight & Initial status \\
\midrule
1 & Budget & ``What budget range do you have in mind for your new home?'' & 0.35 & INCOMPLETE \\
2 & Preferred location & ``In which area or city would you prefer the property to be located?'' & 0.25 & INCOMPLETE \\
3 & Essential requirements & ``Which property characteristics or requirements are essential to you, such as bedrooms, bathrooms, parking, or additional spaces?'' & 0.20 & INCOMPLETE \\
4 & Contact for visit & ``Could you share an email address or telephone number so that we can arrange a joint viewing according to your preferences?'' & 0.20 & INCOMPLETE \\
\bottomrule\end{tabular}\end{table*}

The real-estate vector database was generated from company FAQs and documents covering buyer-support
services, offices and verified contact channels, available properties, locations, prices, bedrooms,
amenities, neighbourhoods, financing, viewing procedures, and reservations. These company facts can
answer a visitor but cannot complete a visitor objective.

Table~\ref{tab:realestate-scenarios} groups the 24 English conversations into six deliberately
challenging families. Standard and multi-goal cases vary the order and density of visitor evidence;
RAG-detour cases insert company-factual questions; validation and repair cases introduce malformed
or revised values; refusal cases test visitor control and later re-engagement; and robustness cases
combine social language, negation, contradictory data, and multilingual input.
\begin{table*}[!t]
\caption{Corner-Case Families in the English Real-Estate Demonstration}
\label{tab:realestate-scenarios}\centering\scriptsize
\begin{tabular}{p{.18\textwidth}cp{.31\textwidth}p{.35\textwidth}}\toprule
Scenario family & Cases & Examples of visitor input & Capability under test \\
\midrule
Standard progression & S01--S04 & Canonical, reverse, location-first, and budget-first disclosure & Goal ordering, extraction outside the prompted order, terminal completion \\
Multi-goal evidence & S05--S08 & Budget plus requirements; three or four goals in one message; factual question plus personal evidence & Simultaneous coordinate completion and evidence isolation \\
RAG detours & S09--S12 & Questions about listings, schools, legal processes, or other company information & Grounded answering, unsupported-information behavior, and resumption without state contamination \\
Validation and repair & S13--S16 & Invalid email or telephone, ambiguous budget, and corrected preference & Validation, clarification, correction, and state commitment \\
Refusal and re-engagement & S17--S20 & Stop request, contact refusal, later company question, or temporary uncertainty & Visitor constraint, NONE, non-fabrication, and voluntary later evidence \\
Robustness & S21--S24 & Social turns, language switching, negation, and contradictory contacts & Irrelevance handling, multilingual extraction, correction, and traceability \\
\bottomrule\end{tabular}
\end{table*}

\subsection{Spanish Professional-Cleaning Demonstration}\label{sec:cleaning-case}
The manager supplied the following intention: ``Identify potential customers interested in
professional cleaning services, collecting the type of client or facility, location, and contact
details needed to prepare personalized advice or a quotation before human intervention''
(``Identificar clientes potenciales interesados en servicios profesionales de limpieza,
recopilando el tipo de cliente o instalación, la ubicación, y los datos de contacto necesarios para
preparar un asesoramiento o presupuesto personalizado antes de la intervención humana''). The
Compiler made the requested cleaning-service type an additional explicit objective and produced
Table~\ref{tab:cleaningcompiler}. The coordinate order is client or facility type, location,
service type, and contact.

\begin{table*}[!t]
\caption{Professional-Cleaning Goals Produced by the LLM Compiler}
\label{tab:cleaningcompiler}\centering\small
\begin{tabular}{clp{.47\textwidth}cc}\toprule
Order & Objective & Canonical question produced by the Compiler & Weight & Initial status \\
\midrule
1 & Client or facility type & ``What type of client or facility requires cleaning services, for example a home, office, hotel, or industrial site?'' & 0.35 & INCOMPLETE \\
2 & Service location & ``In which area or city is the facility requiring the service located?'' & 0.25 & INCOMPLETE \\
3 & Cleaning service type & ``What type of cleaning service are you seeking, for example general cleaning, disinfection, or maintenance?'' & 0.20 & INCOMPLETE \\
4 & Contact & ``Could you provide a telephone number or email address so that we can offer advice or a quotation?'' & 0.20 & INCOMPLETE \\
\bottomrule\end{tabular}
\end{table*}

The cleaning vector database was generated from FAQs compiled by the simulated company. They cover
service areas in Almería, Málaga, and Granada; communities and offices; cleaning frequency;
specialist treatment of marble and other surfaces; windows and facades; hypoallergenic products;
quality control; incident response; professional credentials; and verified contact channels.

Table~\ref{tab:cleaning-scenarios} groups the 10 Spanish conversations by concentrated challenge.
Ordering cases vary sequence and information density; RAG cases interrupt goal collection with
questions about company services; validation and correction cases introduce invalid or revised
values; refusal tests voluntary re-engagement; and the final case tests mixed-language evidence.
\begin{table*}[!t]
\caption{Corner-Case Families in the Spanish Professional-Cleaning Demonstration}
\label{tab:cleaning-scenarios}\centering\scriptsize
\begin{tabular}{p{.18\textwidth}cp{.31\textwidth}p{.35\textwidth}}\toprule
Scenario family & Cases & Examples of visitor input & Capability under test \\
\midrule
Ordering and density & B01--B03 & Canonical order, reverse order, or all four objectives in one message & Out-of-order and multi-coordinate completion \\
Supported RAG detours & B04--B06 & Cleaning frequency, geographic coverage, or marble-floor treatment & Company-grounded answers without state-vector contamination \\
Contact validation & B07 & Invalid email followed by a syntactically valid correction & Validation and correct terminal completion \\
Preference correction & B08 & Location changed explicitly from Almería to Málaga & Value correction while preserving binary completion \\
Refusal and re-engagement & B09 & Contact refusal, quality question, and later voluntary email & Visitor constraint, grounded detour, and resumed completion \\
Multilingual robustness & B10 & English--Spanish language switching within goal-bearing messages & Cross-language semantic extraction \\
\bottomrule\end{tabular}
\end{table*}

\section{Results}\label{sec:results}
Table~\ref{tab:aggregate} summarizes the symbolic performance of the two demonstrations. The
English real-estate condition contains 86 evaluated turns; the Spanish cleaning condition contains
33. For each demonstration, the recorded state vector was interpreted according to the objective
order defined by its own Compiler output.

\begin{table*}[!t]
\caption{Symbolic-State Results for the English and Spanish Demonstrations}
\label{tab:aggregate}\centering\fontsize{7.8}{8.8}\selectfont\setlength{\tabcolsep}{4pt}
\begin{tabular}{lccc}\toprule
Metric & English real estate (24 conversations) & Spanish professional cleaning (10 conversations) & Combined (34 conversations) \\
\midrule
State-transition accuracy (STA) & 73/86 (84.9\%) & 28/33 (84.8\%) & 101/119 (84.9\%) \\
Terminal-state accuracy (TSA) & 22/24 (91.7\%) & 10/10 (100\%) & 32/34 (94.1\%) \\
Monotonicity compliance (MC) & 86/86 (100\%) & 33/33 (100\%) & 119/119 (100\%) \\
Evidence precision (EP) & 83/91 (91.2\%) & 37/40 (92.5\%) & 120/131 (91.6\%) \\
Evidence recall (ER) & 83/94 (88.3\%) & 37/40 (92.5\%) & 120/134 (89.6\%) \\
\bottomrule\end{tabular}
\end{table*}

\subsection{English Real-Estate Demonstration}
Table~\ref{tab:aggregate} shows that 73 of the 86 real-estate turns produced the exact four-coordinate
reference state (STA $=84.9\%$). The system was particularly reliable when factual RAG questions,
social language, negation, or language switching accompanied clear visitor evidence: all 15
robustness turns and 19 of 20 RAG-detour turns matched their reference states. Thus, questions about
listings, processes, or neighbourhood concerns generally did not contaminate the budget, location,
requirements, or contact coordinates.

The 91 and 94 denominators in the evidence rows are completion events, not visitor turns. Several
messages supply two or more objectives simultaneously, so a single turn can contribute multiple
turn--objective pairs. Of 91 completions recorded by GRACE, 83 occurred for the correct objective on
the correct turn (EP $=91.2\%$); the same 83 matched events were found among 94 reference completions
(ER $=88.3\%$). For example, in S02 the message ``I need three bedrooms and a swimming pool'' should
complete the requirements coordinate, but the recorded vector changed the budget coordinate instead.
Conversely, clear multi-goal and RAG-detour turns were usually attributed correctly. These results
explain why the system never reversed a completed coordinate (MC $=100\%$) and still ended with the
correct state in 22 of 24 conversations (TSA $=91.7\%$), despite 13 incorrect intermediate vectors.
Full turn-level evidence appears in Appendix~\ref{app:complete-traces}.

\subsection{Spanish Professional-Cleaning Demonstration}
Table~\ref{tab:aggregate} reports 28 exact states across 33 cleaning turns (STA $=84.8\%$), while all
10 conversations ended in the correct state (TSA $=100\%$) and no coordinate reversed (MC
$=100\%$). Canonical progression, all objectives in one turn, supported RAG interruptions,
invalid-contact repair, and refusal followed by voluntary re-engagement were therefore handled well
at conversation level.

The evidence denominators again count events rather than turns. The 33 visitor turns contain 40
reference completions because some messages satisfy several objectives. GRACE recorded 40 events,
37 of which matched the required objective and turn, yielding EP and ER of $37/40=92.5\%$. The five
non-exact state vectors concentrate in reverse ordering, explicit location correction, and
English--Spanish language switching. The phrase ``office cleaning,'' for example, can be interpreted
as a requested service and can also imply a facility type; on one occasion (B02), GRACE completed
the implied facility-type coordinate earlier than the frozen criterion allowed. This is a failure of criterion-aligned turn
attribution (assigning a completion to the precise objective and visitor turn licensed by the
criterion), not a reversal of stored state. The full Spanish conversations and state vectors appear
in Appendix~\ref{app:cleaning-traces}.

\section{Discussion}\label{sec:discussion}
\subsection{Cross-Domain Findings}
Across both domains and languages, GRACE obtained 84.9\% STA, 91.6\% EP, 89.6\% ER, 100\% MC, and
94.1\% TSA. The experiments support three central claims. First, a goal-oriented layer can coexist with RAG:
visitors received company-grounded answers while the state vector continued to represent only information
about the visitor. Second, compiled objective semantics allowed the same orchestration architecture
to operate in English real estate and Spanish professional cleaning. Third, explicit post-turn
vectors made timing errors visible even when the final conversation appeared successful.

The nearly identical STA values in the two demonstrations do not establish unrestricted transfer.
They do, however, show that the same compiled-goal representation, provenance rule, monotonic update,
and constrained question policy remained operational under different objective meanings, company
corpora, and operating languages. The stronger cleaning TSA suggests that its shorter, more
concentrated scripts allowed displaced intermediate commitments to be corrected before termination.

Corner-case scripts were intentionally used instead of ordinary sales conversations. They
concentrate multi-goal evidence, interruptions, invalid contacts, corrections, refusals, and
language switching into a small evaluation and avoid processing personal data from real prospects. This
design provides strong behavioral coverage but not a population estimate of normal customer use.

\subsection{Error Analysis}
A detailed analysis of the unmatched recorded and reference state vectors reveals two principal
patterns. First, some mismatches occur in dense or semantically overlapping turns. In S13 and S17,
the visitor supplies three objectives in one message but one coordinate remains incomplete; in B02,
the phrase describing ``office cleaning'' expresses a requested service and also implies a facility
type, producing a completion earlier than the frozen criterion permits. Dense evidence is not
intrinsically problematic, however: GRACE correctly completes all four objectives in the single
turn of S07 and all three supplied objectives in the first turn of B07.

Second, some valid evidence is committed one turn late rather than lost. In S05, the budget supplied
with the housing requirements is recorded after the visitor subsequently provides the location. In
S13, the omitted requirements coordinate is completed while the system processes the following
invalid contact, and in B10 the location supplied in the mixed-language opening is completed after
the visitor states the requested cleaning service. These cases suggest that the conversation history
allows GRACE to recover previously missed evidence, improving terminal accuracy, but does not ensure
that every completion is attributed to the exact turn on which its evidence first appears. A tighter
turn-level validator should therefore require all currently supported objectives to be evaluated
before the new state is persisted, while retaining history-based recovery as a safeguard.

\section{Conclusion}\label{sec:conclusion}
This paper has demonstrated an auditable Symbolic--RAG--Generative architecture that combines
company-grounded answering with explicit business-goal progression. GRACE does more than execute a
closed questionnaire and more than answer from a knowledge base: it understands unsolicited and
multi-objective visitor evidence, maintains a visible symbolic state, answers company questions
through a separate RAG component, and selects the next approved objective under a visitor-utility
constraint.

The evaluation is deliberately demanding. It covers 34 conversations, 119 visitor turns, two
business domains, two operating languages, two independently compiled objective sets, and focused
corner cases involving reordered disclosure, dense evidence, RAG interruptions, validation,
correction, refusal, re-engagement, negation, and language switching. GRACE achieved 91.6\% evidence
precision, 100\% monotonicity, and 94.1\% terminal-state accuracy. These results provide compelling
evidence that explicit symbolic control and generative dialogue can be integrated without reducing
the interaction to fixed questions. Equally importantly, the audit record reveals where
turn-level attribution still requires improvement. The contribution is therefore both an
architecture and an empirical demonstration of how its central invariants can be inspected across
domains and languages.

\section{Future Work}\label{sec:future}
Future experiments should compare GRACE directly with the three alternatives in
Table~\ref{tab:position}: a closed-question chatbot, a RAG-only assistant, and a general AI agent.
The comparison should use identical information goals and corpora and report state accuracy,
answer grounding, task completion, visitor burden, latency, computational cost, architectural
complexity, behavioral variance, and human judgments of naturalness and usefulness.

The state model should be extended beyond binary completion to represent uncertain, revised,
revoked, and contradictory preferences. Further work should also evaluate more domains and
languages, larger objective sets, independently authored and naturally occurring conversations,
multiple model families, repeated stochastic runs, and controlled experiments that remove or vary
individual prompt instructions to measure their contribution.

\bibliographystyle{IEEEtran}
\bibliography{references}

\appendices
\onecolumn
\raggedbottom
\section{English Real-Estate Demonstration: Complete Conversation Traces}\label{app:complete-traces}
Tables~\ref{tab:trace-s01}--\ref{tab:trace-s24} report all 24 English-language real-estate
conversations in chronological order. Each row links the visitor message and visible Mentomy
response to the recorded state vector and the evidence-derived reference state after that visitor turn.

\par\medskip\noindent\refstepcounter{table}
\label{tab:trace-s01}
\begin{minipage}{\textwidth}
\begin{center}\textbf{TABLE \thetable}\\[-1pt]
\textsc{S01: Canonical order}\\[2pt]
\fontsize{6.1}{6.7}\selectfont\setlength{\tabcolsep}{2.3pt}
\begin{tabularx}{\textwidth}{cp{.19\textwidth}p{.48\textwidth}cc}
\toprule
Turn & Visitor message & Visible Mentomy response & Recorded & Reference \\
\midrule
1 & My budget is below 350.000 EUR & In which area or city would you prefer your home to be located? & 1000 & 1000 \\
2 & I would like the property to be near Almería. & What specific features or requirements are you looking for in the property, such as the number of bedrooms, bathrooms, parking, or additional spaces? & 1100 & 1100 \\
3 & I need three bedrooms, two bathrooms, and parking. & Could you provide an email address or phone number where we can contact you to arrange a visit? & 1110 & 1110 \\
4 & You can contact me at alex.buyer@example.com. & Thank you. If you shared your contact details, we'll contact you soon. Otherwise, please try asking your question differently or contact Mentomy at info@mentomy.com. & 1111 & 1111 \\
\bottomrule
\end{tabularx}
\end{center}
\end{minipage}

\par\medskip\noindent\refstepcounter{table}
\label{tab:trace-s02}
\begin{minipage}{\textwidth}
\begin{center}\textbf{TABLE \thetable}\\[-1pt]
\textsc{S02: Reverse order}\\[2pt]
\fontsize{6.1}{6.7}\selectfont\setlength{\tabcolsep}{2.3pt}
\begin{tabularx}{\textwidth}{cp{.19\textwidth}p{.48\textwidth}cc}
\toprule
Turn & Visitor message & Visible Mentomy response & Recorded & Reference \\
\midrule
1 & You can contact me at reverse.order@example.com. & What budget range do you have in mind for your new home? & 0001 & 0001 \\
2 & I need three bedrooms and a swimming pool. & In which area or city would you prefer your home to be located? & 1001 & 0011 \\
3 & I want to live in Mojácar. & Which property characteristics or requirements are you looking for, such as bedrooms, bathrooms, parking, or additional spaces? & 1101 & 0111 \\
4 & My maximum budget is 425.000 EUR. & Thank you. If you shared your contact details, we'll contact you soon. Otherwise, please try asking your question differently or contact Mentomy at info@mentomy.com. & 1111 & 1111 \\
\bottomrule
\end{tabularx}
\end{center}
\end{minipage}

\par\medskip\noindent\refstepcounter{table}
\label{tab:trace-s03}
\begin{minipage}{\textwidth}
\begin{center}\textbf{TABLE \thetable}\\[-1pt]
\textsc{S03: Location-first availability}\\[2pt]
\fontsize{6.1}{6.7}\selectfont\setlength{\tabcolsep}{2.3pt}
\begin{tabularx}{\textwidth}{cp{.19\textwidth}p{.48\textwidth}cc}
\toprule
Turn & Visitor message & Visible Mentomy response & Recorded & Reference \\
\midrule
1 & Do you have properties for sale in Vera? & Yes, we have properties for sale in Vera. What budget range do you have in mind for your new home? & 0100 & 0100 \\
2 & I can spend up to 500.000 EUR. & Which property characteristics or requirements are you looking for, such as bedrooms, bathrooms, parking, or additional spaces? & 1100 & 1100 \\
3 & I need four bedrooms and a garden. & Could you share an email address or telephone number so that we can arrange a viewing according to your preferences? & 1110 & 1110 \\
4 & My email is vera.search@example.com. & Thank you. If you shared your contact details, we'll contact you soon. Otherwise, please try asking your question differently or contact Mentomy at info@mentomy.com. & 1111 & 1111 \\
\bottomrule
\end{tabularx}
\end{center}
\end{minipage}

\par\medskip\noindent\refstepcounter{table}
\label{tab:trace-s04}
\begin{minipage}{\textwidth}
\begin{center}\textbf{TABLE \thetable}\\[-1pt]
\textsc{S04: Budget-first property search}\\[2pt]
\fontsize{6.1}{6.7}\selectfont\setlength{\tabcolsep}{2.3pt}
\begin{tabularx}{\textwidth}{cp{.19\textwidth}p{.48\textwidth}cc}
\toprule
Turn & Visitor message & Visible Mentomy response & Recorded & Reference \\
\midrule
1 & I have a budget of 300.000 EUR. Which properties could fit it? & I'm sorry, I don't have enough information to answer your question. In which area or city would you prefer your home to be located? & 1000 & 1000 \\
2 & I prefer somewhere on the coast near Almería. & Which property characteristics or requirements are you looking for, such as bedrooms, bathrooms, parking, or additional spaces? & 1100 & 1100 \\
3 & Two bedrooms and a terrace are essential & Could you share an email address or phone number so we can arrange a visit based on your preferences? & 1110 & 1110 \\
4 & Please use budget.first@example.com. & Thank you. If you shared your contact details, we'll contact you soon. Otherwise, please try asking your question differently or contact Mentomy at info@mentomy.com. & 1111 & 1111 \\
\bottomrule
\end{tabularx}
\end{center}
\end{minipage}

\par\medskip\noindent\refstepcounter{table}
\label{tab:trace-s05}
\begin{minipage}{\textwidth}
\begin{center}\textbf{TABLE \thetable}\\[-1pt]
\textsc{S05: Budget and requirements together}\\[2pt]
\fontsize{6.1}{6.7}\selectfont\setlength{\tabcolsep}{2.3pt}
\begin{tabularx}{\textwidth}{cp{.19\textwidth}p{.48\textwidth}cc}
\toprule
Turn & Visitor message & Visible Mentomy response & Recorded & Reference \\
\midrule
1 & I need three bedrooms and parking, with a maximum budget of 375.000 EUR. & In which area or city would you prefer your home to be located? & 0010 & 1010 \\
2 & I would prefer Mojácar. & Could you please provide an email address or phone number where we can contact you to arrange a visit? & 1110 & 1110 \\
3 & My telephone number is 600000005. & Thank you. If you shared your contact details, we'll contact you soon. Otherwise, please try asking your question differently or contact Mentomy at info@mentomy.com. & 1111 & 1111 \\
\bottomrule
\end{tabularx}
\end{center}
\end{minipage}

\par\medskip\noindent\refstepcounter{table}
\label{tab:trace-s06}
\begin{minipage}{\textwidth}
\begin{center}\textbf{TABLE \thetable}\\[-1pt]
\textsc{S06: Three non-contact objectives together}\\[2pt]
\fontsize{6.1}{6.7}\selectfont\setlength{\tabcolsep}{2.3pt}
\begin{tabularx}{\textwidth}{cp{.19\textwidth}p{.48\textwidth}cc}
\toprule
Turn & Visitor message & Visible Mentomy response & Recorded & Reference \\
\midrule
1 & I am looking in Vera for a four-bedroom house with a pool, below 600.000 EUR. & Could you share an email address or phone number where we can contact you to arrange a viewing? & 1110 & 1110 \\
2 & You can reach me at three.goals@example.com. & Thank you. If you shared your contact details, we'll contact you soon. Otherwise, please try asking your question differently or contact Mentomy at info@mentomy.com. & 1111 & 1111 \\
\bottomrule
\end{tabularx}
\end{center}
\end{minipage}

\par\medskip\noindent\refstepcounter{table}
\label{tab:trace-s07}
\begin{minipage}{\textwidth}
\begin{center}\textbf{TABLE \thetable}\\[-1pt]
\textsc{S07: All objectives in one message}\\[2pt]
\fontsize{6.1}{6.7}\selectfont\setlength{\tabcolsep}{2.3pt}
\begin{tabularx}{\textwidth}{cp{.19\textwidth}p{.48\textwidth}cc}
\toprule
Turn & Visitor message & Visible Mentomy response & Recorded & Reference \\
\midrule
1 & I want a three-bedroom apartment in Mojácar for no more than 325.000 EUR. Contact me at all.at.once@example.com. & Thank you. If you shared your contact details, we'll contact you soon. Otherwise, please try asking your question differently or contact Mentomy at info@mentomy.com. & 1111 & 1111 \\
\bottomrule
\end{tabularx}
\end{center}
\end{minipage}

\par\medskip\noindent\refstepcounter{table}
\label{tab:trace-s08}
\begin{minipage}{\textwidth}
\begin{center}\textbf{TABLE \thetable}\\[-1pt]
\textsc{S08: Two goals plus a factual question}\\[2pt]
\fontsize{6.1}{6.7}\selectfont\setlength{\tabcolsep}{2.3pt}
\begin{tabularx}{\textwidth}{cp{.19\textwidth}p{.48\textwidth}cc}
\toprule
Turn & Visitor message & Visible Mentomy response & Recorded & Reference \\
\midrule
1 & My budget is 450.000 EUR and I prefer Vera. Do you have properties there with parking? & Yes, we have properties in Vera within your EUR 450,000 budget that include parking. What specific features or requirements would you like in the home, such as the number of bedrooms or bathrooms? & 1100 & 1100 \\
2 & I need at least three bedrooms; parking is also essential. & Could you share an email address or phone number where we can contact you to schedule a visit? & 1110 & 1110 \\
3 & My email is parking.search@example.com. & Thank you. If you shared your contact details, we'll contact you soon. Otherwise, please try asking your question differently or contact Mentomy at info@mentomy.com. & 1111 & 1111 \\
\bottomrule
\end{tabularx}
\end{center}
\end{minipage}

\par\medskip\noindent\refstepcounter{table}
\label{tab:trace-s09}
\begin{minipage}{\textwidth}
\begin{center}\textbf{TABLE \thetable}\\[-1pt]
\textsc{S09: Supported availability detour}\\[2pt]
\fontsize{6.1}{6.7}\selectfont\setlength{\tabcolsep}{2.3pt}
\begin{tabularx}{\textwidth}{cp{.19\textwidth}p{.48\textwidth}cc}
\toprule
Turn & Visitor message & Visible Mentomy response & Recorded & Reference \\
\midrule
1 & I want to buy near Almería. & Yes, we have properties for sale near Almería within your EUR 450,000 budget that include parking. What budget range do you have in mind for your new home? & 0100 & 0100 \\
2 & My budget is below 350.000 EUR. & Which property characteristics or requirements are you looking for, such as bedrooms, bathrooms, parking, or additional spaces? & 1100 & 1100 \\
3 & Do you have houses with a swimming pool? & Yes, we have houses with swimming pools for sale in Vera. Would you like to schedule a visit to see properties with swimming pools in Vera? & 1100 & 1100 \\
4 & I need three bedrooms and a pool. & Could you share an email address or phone number so we can arrange a visit to see properties that match your needs? & 1110 & 1110 \\
5 & Contact me at supported.detour@example.com. & Thank you. If you shared your contact details, we'll contact you soon. Otherwise, please try asking your question differently or contact Mentomy at info@mentomy.com. & 1111 & 1111 \\
\bottomrule
\end{tabularx}
\end{center}
\end{minipage}

\par\medskip\noindent\refstepcounter{table}
\label{tab:trace-s10}
\begin{minipage}{\textwidth}
\begin{center}\textbf{TABLE \thetable}\\[-1pt]
\textsc{S10: Unsupported schools question}\\[2pt]
\fontsize{6.1}{6.7}\selectfont\setlength{\tabcolsep}{2.3pt}
\begin{tabularx}{\textwidth}{cp{.19\textwidth}p{.48\textwidth}cc}
\toprule
Turn & Visitor message & Visible Mentomy response & Recorded & Reference \\
\midrule
1 & I am looking for a home in Mojácar. & What budget range do you have in mind for your new home? & 0100 & 0100 \\
2 & Are there international schools near the available properties? & I'm sorry, I don't have enough information to answer your question. Which property characteristics or requirements are you looking for, such as bedrooms, bathrooms, parking, or additional spaces? & 0100 & 0100 \\
3 & I need three bedrooms and two bathrooms. & What budget range do you have in mind for your new home? & 0110 & 0110 \\
4 & My maximum budget is 400.000 EUR. & Could you share an email address or phone number where we can contact you to arrange a personalized visit? & 1110 & 1110 \\
5 & Use schools.test@example.com. & Thank you. If you shared your contact details, we'll contact you soon. Otherwise, please try asking your question differently or contact Mentomy at info@mentomy.com. & 1111 & 1111 \\
\bottomrule
\end{tabularx}
\end{center}
\end{minipage}

\par\medskip\noindent\refstepcounter{table}
\label{tab:trace-s11}
\begin{minipage}{\textwidth}
\begin{center}\textbf{TABLE \thetable}\\[-1pt]
\textsc{S11: Supported legal-process interruption}\\[2pt]
\fontsize{6.1}{6.7}\selectfont\setlength{\tabcolsep}{2.3pt}
\begin{tabularx}{\textwidth}{cp{.19\textwidth}p{.48\textwidth}cc}
\toprule
Turn & Visitor message & Visible Mentomy response & Recorded & Reference \\
\midrule
1 & I want a property in Vera and can spend up to 550.000 EUR. & Which property characteristics or requirements are you looking for, such as bedrooms, bathrooms, parking, or additional spaces? & 1000 & 1100 \\
2 & As a foreign buyer, what documentation do I need to purchase a home in Spain? & As a foreign buyer, you need:- NIE (Foreigner Identification Number).- Spanish bank account.- Reservation contract (contrato de arras) before the final purchase. Which property characteristics or requirements are you looking for, such as bedrooms, bathrooms, parking, or additional spaces? & 1100 & 1100 \\
3 & I need four bedrooms, parking, and outdoor space. & Could you provide an email address or phone number where we can contact you to arrange a viewing? & 1110 & 1110 \\
4 & My email is foreign.buyer@example.com. & Thank you. If you shared your contact details, we'll contact you soon. Otherwise, please try asking your question differently or contact Mentomy at info@mentomy.com. & 1111 & 1111 \\
\bottomrule
\end{tabularx}
\end{center}
\end{minipage}

\par\medskip\noindent\refstepcounter{table}
\label{tab:trace-s12}
\begin{minipage}{\textwidth}
\begin{center}\textbf{TABLE \thetable}\\[-1pt]
\textsc{S12: Two consecutive factual detours}\\[2pt]
\fontsize{6.1}{6.7}\selectfont\setlength{\tabcolsep}{2.3pt}
\begin{tabularx}{\textwidth}{cp{.19\textwidth}p{.48\textwidth}cc}
\toprule
Turn & Visitor message & Visible Mentomy response & Recorded & Reference \\
\midrule
1 & I am interested in Mojácar. & What budget range do you have in mind for your new home? & 0100 & 0100 \\
2 & What services do you provide to buyers? & We provide three main services to buyers:- Property sales (homes, chalets, fincas).- Tourist rental management.- Personalized real estate advice, including profitability analysis for investors. & 0100 & 0100 \\
3 & How can I arrange a property viewing? & I'm sorry, I don't have enough information to answer your question. Which property characteristics or requirements are you looking for, such as bedrooms, bathrooms, parking, or additional spaces? & 0100 & 0100 \\
4 & I need three bedrooms and a terrace. & What budget range do you have in mind for your new home? & 0110 & 0110 \\
5 & My budget is 475.000 EUR. & Could you share an email address or telephone number so that we can arrange a viewing according to your preferences? & 1110 & 1110 \\
6 & Contact me at two.detours@example.com. & Thank you. If you shared your contact details, we'll contact you soon. Otherwise, please try asking your question differently or contact Mentomy at info@mentomy.com. & 1111 & 1111 \\
\bottomrule
\end{tabularx}
\end{center}
\end{minipage}

\par\medskip\noindent\refstepcounter{table}
\label{tab:trace-s13}
\begin{minipage}{\textwidth}
\begin{center}\textbf{TABLE \thetable}\\[-1pt]
\textsc{S13: Invalid email and correction}\\[2pt]
\fontsize{6.1}{6.7}\selectfont\setlength{\tabcolsep}{2.3pt}
\begin{tabularx}{\textwidth}{cp{.19\textwidth}p{.48\textwidth}cc}
\toprule
Turn & Visitor message & Visible Mentomy response & Recorded & Reference \\
\midrule
1 & I can spend 350.000 EUR, prefer Almería, and need three bedrooms. & Which property characteristics or requirements are you looking for, such as bedrooms, bathrooms, parking, or additional spaces? & 1100 & 1110 \\
2 & My email is buyer@invalid. & Could you please verify the email address you provided? It appears to be invalid. & 1110 & 1110 \\
3 & Correction: my email is buyer.valid@example.com. & Thank you. If you shared your contact details, we'll contact you soon. Otherwise, please try asking your question differently or contact Mentomy at info@mentomy.com. & 1111 & 1111 \\
\bottomrule
\end{tabularx}
\end{center}
\end{minipage}

\par\medskip\noindent\refstepcounter{table}
\label{tab:trace-s14}
\begin{minipage}{\textwidth}
\begin{center}\textbf{TABLE \thetable}\\[-1pt]
\textsc{S14: Invalid telephone and correction}\\[2pt]
\fontsize{6.1}{6.7}\selectfont\setlength{\tabcolsep}{2.3pt}
\begin{tabularx}{\textwidth}{cp{.19\textwidth}p{.48\textwidth}cc}
\toprule
Turn & Visitor message & Visible Mentomy response & Recorded & Reference \\
\midrule
1 & I want a two-bedroom apartment in Vera below 300.000 EUR. & Which other specific characteristics or requirements are you looking for in the property? & 1100 & 1110 \\
2 & My telephone number is 12. & Could you please provide a valid phone number or email address so we can contact you? & 1100 & 1110 \\
3 & The complete number is 600000014. & Which property characteristics or requirements are you looking for, such as bedrooms, bathrooms, parking, or additional spaces? & 1101 & 1111 \\
4$^a$ & End of the conversation & Thank you. If you shared your contact details, we'll contact you soon. Otherwise, please try asking your question differently or contact Mentomy at info@mentomy.com. & 1101 & -- \\
\bottomrule
\end{tabularx}
\end{center}
\end{minipage}

\par\medskip\noindent\refstepcounter{table}
\label{tab:trace-s15}
\begin{minipage}{\textwidth}
\begin{center}\textbf{TABLE \thetable}\\[-1pt]
\textsc{S15: Ambiguous budget and clarification}\\[2pt]
\fontsize{6.1}{6.7}\selectfont\setlength{\tabcolsep}{2.3pt}
\begin{tabularx}{\textwidth}{cp{.19\textwidth}p{.48\textwidth}cc}
\toprule
Turn & Visitor message & Visible Mentomy response & Recorded & Reference \\
\midrule
1 & I am looking in Mojácar for a three-bedroom apartment. & What budget range do you have in mind for your new home? & 0100 & 0110 \\
2 & My budget is around three hundred. & Could you clarify your budget in euros or specify a more precise range? & 0110 & 0110 \\
3 & My budget is around three hundred thousand euros. & Could you share an email address or phone number where we can contact you to arrange a visit? & 1110 & 1110 \\
4 & Use ambiguous.budget@example.com. & Thank you. If you shared your contact details, we'll contact you soon. Otherwise, please try asking your question differently or contact Mentomy at info@mentomy.com. & 1111 & 1111 \\
\bottomrule
\end{tabularx}
\end{center}
\end{minipage}

\par\medskip\noindent\refstepcounter{table}
\label{tab:trace-s16}
\begin{minipage}{\textwidth}
\begin{center}\textbf{TABLE \thetable}\\[-1pt]
\textsc{S16: Explicit location correction}\\[2pt]
\fontsize{6.1}{6.7}\selectfont\setlength{\tabcolsep}{2.3pt}
\begin{tabularx}{\textwidth}{cp{.19\textwidth}p{.48\textwidth}cc}
\toprule
Turn & Visitor message & Visible Mentomy response & Recorded & Reference \\
\midrule
1 & I prefer Almería and my budget is 500.000 EUR. & Which property characteristics or requirements are you looking for, such as bedrooms, bathrooms, parking, or additional spaces? & 1100 & 1100 \\
2 & Actually, replace Almería with Vera as my preferred location. & Which property characteristics or requirements are you looking for, such as bedrooms, bathrooms, parking, or additional spaces? & 1100 & 1100 \\
3 & I need four bedrooms and a garage. & Could you share an email address or phone number where we can contact you to schedule a visit? & 1110 & 1110 \\
4 & My email is corrected.location@example.com. & Thank you. If you shared your contact details, we'll contact you soon. Otherwise, please try asking your question differently or contact Mentomy at info@mentomy.com. & 1111 & 1111 \\
\bottomrule
\end{tabularx}
\end{center}
\end{minipage}

\par\medskip\noindent\refstepcounter{table}
\label{tab:trace-s17}
\begin{minipage}{\textwidth}
\begin{center}\textbf{TABLE \thetable}\\[-1pt]
\textsc{S17: Contact refusal and explicit stop}\\[2pt]
\fontsize{6.1}{6.7}\selectfont\setlength{\tabcolsep}{2.3pt}
\begin{tabularx}{\textwidth}{cp{.19\textwidth}p{.48\textwidth}cc}
\toprule
Turn & Visitor message & Visible Mentomy response & Recorded & Reference \\
\midrule
1 & I want a three-bedroom house in Vera below 450.000 EUR. & Which additional features, such as bathrooms, parking, or other spaces, are important to you? & 1100 & 1110 \\
2 & I will not provide contact details. Please stop asking me questions. & Thank you. If you shared your contact details, we'll contact you soon. Otherwise, please try asking your question differently or contact Mentomy at info@mentomy.com. & 1110 & 1110 \\
\bottomrule
\end{tabularx}
\end{center}
\end{minipage}

\par\medskip\noindent\refstepcounter{table}
\label{tab:trace-s18}
\begin{minipage}{\textwidth}
\begin{center}\textbf{TABLE \thetable}\\[-1pt]
\textsc{S18: Contact refusal and re-engagement}\\[2pt]
\fontsize{6.1}{6.7}\selectfont\setlength{\tabcolsep}{2.3pt}
\begin{tabularx}{\textwidth}{cp{.19\textwidth}p{.48\textwidth}cc}
\toprule
Turn & Visitor message & Visible Mentomy response & Recorded & Reference \\
\midrule
1 & I want a three-bedroom apartment in Mojácar below 400.000 EUR. & Which property characteristics or requirements are you looking for, such as bedrooms, bathrooms, parking, or additional spaces? & 1100 & 1110 \\
2 & I do not want to provide contact information. & Thank you. If you shared your contact details, we'll contact you soon. Otherwise, please try asking your question differently or contact Mentomy at info@mentomy.com. & 1110 & 1110 \\
3 & Do you help buyers arrange property viewings? & Thank you. If you shared your contact details, we'll contact you soon. Otherwise, please try asking your question differently or contact Mentomy at info@mentomy.com. & 1110 & 1110 \\
4 & I have reconsidered. You can use reconsidered@example.com. & Thank you. If you shared your contact details, we'll contact you soon. Otherwise, please try asking your question differently or contact Mentomy at info@mentomy.com. & 1111 & 1111 \\
\bottomrule
\end{tabularx}
\end{center}
\end{minipage}

\par\medskip\noindent\refstepcounter{table}
\label{tab:trace-s19}
\begin{minipage}{\textwidth}
\begin{center}\textbf{TABLE \thetable}\\[-1pt]
\textsc{S19: Budget-specific refusal}\\[2pt]
\fontsize{6.1}{6.7}\selectfont\setlength{\tabcolsep}{2.3pt}
\begin{tabularx}{\textwidth}{cp{.19\textwidth}p{.48\textwidth}cc}
\toprule
Turn & Visitor message & Visible Mentomy response & Recorded & Reference \\
\midrule
1 & I want a two-bedroom apartment in Almería. You can contact me at no.budget@example.com. & [WARNING] No grounded information found in your knowledge base for this question. To avoid generating unverified or speculative content, Mentomy does not answer questions that cannot be supported by your selected documents. Please refine your question or ensure the relevant documents are available. What budget range do you have in mind for your new home? & 0010 & 0111 \\
2 & I prefer not to disclose my budget. & In which area or city would you prefer your home to be located? & 0010 & 0111 \\
3$^a$ & Bye & Thank you. If you shared your contact details, we'll contact you soon. Otherwise, please try asking your question differently or contact Mentomy at info@mentomy.com. & 0010 & -- \\
\bottomrule
\end{tabularx}
\end{center}
\end{minipage}

\par\medskip\noindent\refstepcounter{table}
\label{tab:trace-s20}
\begin{minipage}{\textwidth}
\begin{center}\textbf{TABLE \thetable}\\[-1pt]
\textsc{S20: Temporary uncertainty and later evidence}\\[2pt]
\fontsize{6.1}{6.7}\selectfont\setlength{\tabcolsep}{2.3pt}
\begin{tabularx}{\textwidth}{cp{.19\textwidth}p{.48\textwidth}cc}
\toprule
Turn & Visitor message & Visible Mentomy response & Recorded & Reference \\
\midrule
1 & I want a home in Vera with three bedrooms and parking. & What budget range do you have in mind for your new home? & 0110 & 0110 \\
2 & I do not know my budget yet. & Thank you. If you shared your contact details, we'll contact you soon. Otherwise, please try asking your question differently or contact Mentomy at info@mentomy.com. & 0110 & 0110 \\
3 & Please use temporary.uncertainty@example.com. & Thank you. If you shared your contact details, we'll contact you soon. Otherwise, please try asking your question differently or contact Mentomy at info@mentomy.com. & 0111 & 0111 \\
4 & I have checked: my maximum budget is 525.000 EUR. & Thank you. If you shared your contact details, we'll contact you soon. Otherwise, please try asking your question differently or contact Mentomy at info@mentomy.com. & 1111 & 1111 \\
\bottomrule
\end{tabularx}
\end{center}
\end{minipage}

\par\medskip\noindent\refstepcounter{table}
\label{tab:trace-s21}
\begin{minipage}{\textwidth}
\begin{center}\textbf{TABLE \thetable}\\[-1pt]
\textsc{S21: Irrelevant social turns}\\[2pt]
\fontsize{6.1}{6.7}\selectfont\setlength{\tabcolsep}{2.3pt}
\begin{tabularx}{\textwidth}{cp{.19\textwidth}p{.48\textwidth}cc}
\toprule
Turn & Visitor message & Visible Mentomy response & Recorded & Reference \\
\midrule
1 & Hello, I am Peter. & What budget range do you have in mind for your new home? & 0000 & 0000 \\
2 & It is a beautiful day today. & What budget range do you have in mind for your new home? & 0000 & 0000 \\
3 & I want to buy in Vera. & [WARNING] No grounded information found in your knowledge base for this question. To avoid generating unverified or speculative content, Mentomy does not answer questions that cannot be supported by your selected documents. Please refine your question or ensure the relevant documents are available. What budget range do you have in mind for your new home? & 0100 & 0100 \\
4 & My budget is 450.000 EUR and I need three bedrooms. & Could you share an email address or telephone number so that we can arrange a viewing according to your preferences? & 1110 & 1110 \\
5 & My email is social.turns@example.com. & Thank you. If you shared your contact details, we'll contact you soon. Otherwise, please try asking your question differently or contact Mentomy at info@mentomy.com. & 1111 & 1111 \\
\bottomrule
\end{tabularx}
\end{center}
\end{minipage}

\par\medskip\noindent\refstepcounter{table}
\label{tab:trace-s22}
\begin{minipage}{\textwidth}
\begin{center}\textbf{TABLE \thetable}\\[-1pt]
\textsc{S22: English-Spanish language switching}\\[2pt]
\fontsize{6.1}{6.7}\selectfont\setlength{\tabcolsep}{2.3pt}
\begin{tabularx}{\textwidth}{cp{.19\textwidth}p{.48\textwidth}cc}
\toprule
Turn & Visitor message & Visible Mentomy response & Recorded & Reference \\
\midrule
1 & I want to buy in Mojácar, near the beach. & Mojácar, particularly the Macenas urbanization, is recommended for buying near the beach. Which property characteristics or requirements are you looking for, such as bedrooms, bathrooms, parking, or additional spaces? & 0100 & 0100 \\
2 & I need three bedrooms and parking. & What budget range do you have in mind for your new home? & 0110 & 0110 \\
3 & My maximum budget is 375.000 EUR. & Could you share an email address or telephone number so that we can arrange a viewing according to your preferences? & 1110 & 1110 \\
4 & You can contact me at code.switch@example.com. & Thank you. If you shared your contact details, we'll contact you soon. Otherwise, please try asking your question differently or contact Mentomy at info@mentomy.com. & 1111 & 1111 \\
\bottomrule
\end{tabularx}
\end{center}
\end{minipage}

\par\medskip\noindent\refstepcounter{table}
\label{tab:trace-s23}
\begin{minipage}{\textwidth}
\begin{center}\textbf{TABLE \thetable}\\[-1pt]
\textsc{S23: Negation and corrected requirement}\\[2pt]
\fontsize{6.1}{6.7}\selectfont\setlength{\tabcolsep}{2.3pt}
\begin{tabularx}{\textwidth}{cp{.19\textwidth}p{.48\textwidth}cc}
\toprule
Turn & Visitor message & Visible Mentomy response & Recorded & Reference \\
\midrule
1 & I am looking in Almería with a budget of 425.000 EUR. & Which property characteristics or requirements are you looking for, such as bedrooms, bathrooms, parking, or additional spaces? & 1100 & 1100 \\
2 & I do not need parking. What I do need is three bedrooms and two bathrooms. & Could you share an email address or telephone number so that we can arrange a viewing according to your preferences? & 1110 & 1110 \\
3 & Contact me at negation.test@example.com. & Thank you. If you shared your contact details, we'll contact you soon. Otherwise, please try asking your question differently or contact Mentomy at info@mentomy.com. & 1111 & 1111 \\
\bottomrule
\end{tabularx}
\end{center}
\end{minipage}

\par\medskip\noindent\refstepcounter{table}
\label{tab:trace-s24}
\begin{minipage}{\textwidth}
\begin{center}\textbf{TABLE \thetable}\\[-1pt]
\textsc{S24: Contradictory contact and explicit selection}\\[2pt]
\fontsize{6.1}{6.7}\selectfont\setlength{\tabcolsep}{2.3pt}
\begin{tabularx}{\textwidth}{cp{.19\textwidth}p{.48\textwidth}cc}
\toprule
Turn & Visitor message & Visible Mentomy response & Recorded & Reference \\
\midrule
1 & I need a three-bedroom house in Vera below 500.000 EUR. & Could you share an email address or telephone number so that we can arrange a viewing according to your preferences? & 1110 & 1110 \\
2 & My contacts are first.contact@example.com and second.contact@example.com. & Thank you. If you shared your contact details, we'll contact you soon. Otherwise, please try asking your question differently or contact Mentomy at info@mentomy.com. & 1111 & 1111 \\
3 & Please use only second.contact@example.com and disregard the first address. & Thank you. If you shared your contact details, we'll contact you soon. Otherwise, please try asking your question differently or contact Mentomy at info@mentomy.com. & 1111 & 1111 \\
\bottomrule
\end{tabularx}
\end{center}
\end{minipage}

\section{Spanish Professional-Cleaning Demonstration: Complete Conversation Traces}\label{app:cleaning-traces}
Tables~\ref{tab:clean-b01}--\ref{tab:clean-b10} report all 10 Spanish-language
professional-cleaning conversations in chronological order. Each row links the visitor message and
visible Mentomy response to the recorded state vector and the evidence-derived reference state after that visitor turn.

\par\smallskip\noindent\refstepcounter{table}
\label{tab:clean-b01}
\begin{minipage}{\textwidth}
\begin{center}\textbf{TABLE \thetable}\\[-1pt]
\textsc{B01: Canonical progression}\\[2pt]
\fontsize{5.8}{6.2}\selectfont\setlength{\tabcolsep}{2.1pt}
\begin{tabularx}{\textwidth}{cp{.19\textwidth}p{.48\textwidth}cc}
\toprule
Turn & Visitor message & Visible Mentomy response & Recorded & Reference \\
\midrule
1 & El tipo de instalación que necesita servicios de limpieza es una oficina. & Sí, ofrecemos limpieza de oficinas en Málaga adaptada a vuestras necesidades. ¿En qué zona o ciudad se encuentra su oficina? & 1000 & 1000 \\
2 & La oficina está situada en Almería. & ¿Qué tipo de servicios de limpieza está buscando? & 1100 & 1100 \\
3 & Busco un servicio de limpieza general de oficinas. & Sí, ofrecemos limpieza de oficinas en Málaga, Almería y Granada adaptada al horario de la empresa. ¿Podría facilitarnos un teléfono o correo electrónico para contactarle? & 1110 & 1110 \\
4 & Mi correo electrónico de contacto es balanced.b01@example.com. & Gracias. Si nos has dejado tus datos, nos pondremos en contacto contigo pronto. Si no, prueba a reformular tu pregunta o contacta con Mentomy en info@mentomy.com. & 1111 & 1111 \\
\bottomrule
\end{tabularx}
\end{center}
\end{minipage}

\par\smallskip\noindent\refstepcounter{table}
\label{tab:clean-b02}
\begin{minipage}{\textwidth}
\begin{center}\textbf{TABLE \thetable}\\[-1pt]
\textsc{B02: Reverse progression}\\[2pt]
\fontsize{5.8}{6.2}\selectfont\setlength{\tabcolsep}{2.1pt}
\begin{tabularx}{\textwidth}{cp{.19\textwidth}p{.48\textwidth}cc}
\toprule
Turn & Visitor message & Visible Mentomy response & Recorded & Reference \\
\midrule
1 & Mi correo electrónico de contacto es balanced.b02@example.com. & ¿Qué tipo de cliente o instalación necesita servicios de limpieza? Por ejemplo, hogar, oficina, hotel, industria, etc. & 0001 & 0001 \\
2 & Busco un servicio de limpieza general de oficinas. & Sí, ofrecemos limpieza de oficinas en Almería, Málaga y Granada adaptada al horario de la empresa. ¿En qué zona o ciudad se encuentra su oficina? & 1011 & 0011 \\
3 & La instalación está situada en Málaga. & Gracias. Si nos has dejado tus datos, nos pondremos en contacto contigo pronto. Si no, prueba a reformular tu pregunta o contacta con Mentomy en info@mentomy.com. & 1111 & 0111 \\
4 & El tipo de instalación que necesita el servicio es una oficina. & Gracias. Si nos has dejado tus datos, nos pondremos en contacto contigo pronto. Si no, prueba a reformular tu pregunta o contacta con Mentomy en info@mentomy.com. & 1111 & 1111 \\
\bottomrule
\end{tabularx}
\end{center}
\end{minipage}

\par\smallskip\noindent\refstepcounter{table}
\label{tab:clean-b03}
\begin{minipage}{\textwidth}
\begin{center}\textbf{TABLE \thetable}\\[-1pt]
\textsc{B03: All objectives together}\\[2pt]
\fontsize{5.8}{6.2}\selectfont\setlength{\tabcolsep}{2.1pt}
\begin{tabularx}{\textwidth}{cp{.19\textwidth}p{.48\textwidth}cc}
\toprule
Turn & Visitor message & Visible Mentomy response & Recorded & Reference \\
\midrule
1 & El tipo de instalación es una oficina situada en Granada, busco un servicio de limpieza general de oficinas y mi correo electrónico de contacto es balanced.b03@example.com. & Sí, ofrecemos limpieza de oficinas en Granada adaptada al horario de la empresa. & 1111 & 1111 \\
\bottomrule
\end{tabularx}
\end{center}
\end{minipage}

\par\smallskip\noindent\refstepcounter{table}
\label{tab:clean-b04}
\begin{minipage}{\textwidth}
\begin{center}\textbf{TABLE \thetable}\\[-1pt]
\textsc{B04: Frequency RAG interruption}\\[2pt]
\fontsize{5.8}{6.2}\selectfont\setlength{\tabcolsep}{2.1pt}
\begin{tabularx}{\textwidth}{cp{.19\textwidth}p{.48\textwidth}cc}
\toprule
Turn & Visitor message & Visible Mentomy response & Recorded & Reference \\
\midrule
1 & El tipo de cliente es una comunidad de propietarios situada en Almería. & Sí, ofrecemos servicio de limpieza para comunidades de propietarios en Almería. ¿Qué tipo de servicios de limpieza está buscando para su comunidad de propietarios? & 1100 & 1100 \\
2 & Antes de concretar el servicio, ¿con qué frecuencia soléis limpiar las zonas comunes? & Lo habitual es entre 2 y 5 días por semana para portal y escaleras, según el tipo de comunidad y lo acordado. & 1100 & 1100 \\
3 & Busco un servicio de limpieza general del portal y las escaleras cuatro días por semana. & Sí, podemos limpiar el portal y las escaleras cuatro días por semana en su comunidad de Almería. ¿Podría facilitarnos un teléfono o correo electrónico para contactarle y enviarle un presupuesto personalizado? & 1110 & 1110 \\
4 & Mi teléfono de contacto es 600200004. & Gracias. Si nos has dejado tus datos, nos pondremos en contacto contigo pronto. Si no, prueba a reformular tu pregunta o contacta con Mentomy en info@mentomy.com. & 1111 & 1111 \\
\bottomrule
\end{tabularx}
\end{center}
\end{minipage}

\par\smallskip\noindent\refstepcounter{table}
\label{tab:clean-b05}
\begin{minipage}{\textwidth}
\begin{center}\textbf{TABLE \thetable}\\[-1pt]
\textsc{B05: Service-area RAG question}\\[2pt]
\fontsize{5.8}{6.2}\selectfont\setlength{\tabcolsep}{2.1pt}
\begin{tabularx}{\textwidth}{cp{.19\textwidth}p{.48\textwidth}cc}
\toprule
Turn & Visitor message & Visible Mentomy response & Recorded & Reference \\
\midrule
1 & El tipo de cliente es una empresa y sus oficinas están en Granada. ¿Dais servicio allí? & Sí, damos servicio de limpieza en Granada para empresas. ¿Qué tipo de servicios de limpieza necesita para sus oficinas? & 1100 & 1100 \\
2 & Busco un servicio de limpieza general de las oficinas. & Sí, ofrecemos limpieza de oficinas en Granada adaptada al horario de la empresa. ¿Podría facilitarnos un teléfono o correo electrónico para contactarle? & 1110 & 1110 \\
3 & Mi correo electrónico de contacto es balanced.b05@example.com. & Gracias. Si nos has dejado tus datos, nos pondremos en contacto contigo pronto. Si no, prueba a reformular tu pregunta o contacta con Mentomy en info@mentomy.com. & 1111 & 1111 \\
\bottomrule
\end{tabularx}
\end{center}
\end{minipage}

\par\smallskip\noindent\refstepcounter{table}
\label{tab:clean-b06}
\begin{minipage}{\textwidth}
\begin{center}\textbf{TABLE \thetable}\\[-1pt]
\textsc{B06: Specialist-service RAG question}\\[2pt]
\fontsize{5.8}{6.2}\selectfont\setlength{\tabcolsep}{2.1pt}
\begin{tabularx}{\textwidth}{cp{.19\textwidth}p{.48\textwidth}cc}
\toprule
Turn & Visitor message & Visible Mentomy response & Recorded & Reference \\
\midrule
1 & El cliente es una comunidad de propietarios y el edificio está situado en Málaga. & Sí, ofrecemos servicio de limpieza para comunidades de propietarios en Málaga. ¿Qué tipo de servicios de limpieza necesita para su comunidad de propietarios? & 1100 & 1100 \\
2 & Busco un servicio de limpieza y tratamiento del suelo de mármol del portal. ¿Podéis realizarlo sin dañar el material? & Sí, podemos realizar el tratamiento del suelo de mármol del portal sin dañar el material. Para poder enviarle un presupuesto personalizado, ¿podría facilitarnos un teléfono o correo electrónico donde podamos contactarle? & 1110 & 1110 \\
3 & Mi teléfono de contacto es 600200006. & Gracias. Si nos has dejado tus datos, nos pondremos en contacto contigo pronto. Si no, prueba a reformular tu pregunta o contacta con Mentomy en info@mentomy.com. & 1111 & 1111 \\
\bottomrule
\end{tabularx}
\end{center}
\end{minipage}

\par\smallskip\noindent\refstepcounter{table}
\label{tab:clean-b07}
\begin{minipage}{\textwidth}
\begin{center}\textbf{TABLE \thetable}\\[-1pt]
\textsc{B07: Invalid email and correction}\\[2pt]
\fontsize{5.8}{6.2}\selectfont\setlength{\tabcolsep}{2.1pt}
\begin{tabularx}{\textwidth}{cp{.19\textwidth}p{.48\textwidth}cc}
\toprule
Turn & Visitor message & Visible Mentomy response & Recorded & Reference \\
\midrule
1 & El tipo de instalación es una oficina situada en Almería y busco un servicio de limpieza general de oficinas. & Sí, ofrecemos servicio de limpieza general para oficinas en Almería. ¿Podría indicarnos un teléfono o correo electrónico donde podamos contactarlo para ofrecerle un asesoramiento o presupuesto? & 1110 & 1110 \\
2 & Mi correo electrónico es oficina@invalid. & Por favor, verifique el correo electrónico que nos proporcionó, ya que parece no ser válido. ¿Podría indicarnos uno correcto? & 1110 & 1110 \\
3 & Corrijo el dato: mi correo electrónico de contacto es balanced.b07@example.com. & Gracias. Si nos has dejado tus datos, nos pondremos en contacto contigo pronto. Si no, prueba a reformular tu pregunta o contacta con Mentomy en info@mentomy.com. & 1111 & 1111 \\
\bottomrule
\end{tabularx}
\end{center}
\end{minipage}

\par\smallskip\noindent\refstepcounter{table}
\label{tab:clean-b08}
\begin{minipage}{\textwidth}
\begin{center}\textbf{TABLE \thetable}\\[-1pt]
\textsc{B08: Location correction}\\[2pt]
\fontsize{5.8}{6.2}\selectfont\setlength{\tabcolsep}{2.1pt}
\begin{tabularx}{\textwidth}{cp{.19\textwidth}p{.48\textwidth}cc}
\toprule
Turn & Visitor message & Visible Mentomy response & Recorded & Reference \\
\midrule
1 & El tipo de instalación es una oficina situada en Almería. & ¿Qué tipo de servicios de limpieza está buscando? Por ejemplo, limpieza general, desinfección, mantenimiento, etc. & 0100 & 1100 \\
2 & Corrijo la ubicación: la oficina está en Málaga, no en Almería. & ¿Qué tipo de servicios de limpieza está buscando? Por ejemplo, limpieza general, desinfección, mantenimiento, etc. & 0100 & 1100 \\
3 & Busco un servicio de limpieza general de oficinas. & Sí, ofrecemos servicio de limpieza general para oficinas en Almería, Málaga y Granada. ¿Podría indicarnos un teléfono o correo electrónico donde podamos contactarlo para ofrecerle un asesoramiento o presupuesto? & 1110 & 1110 \\
4 & Mi teléfono de contacto es 600200008. & Sí, podemos contactarle al teléfono 600.200.008. & 1111 & 1111 \\
\bottomrule
\end{tabularx}
\end{center}
\end{minipage}

\par\smallskip\noindent\refstepcounter{table}
\label{tab:clean-b09}
\begin{minipage}{\textwidth}
\begin{center}\textbf{TABLE \thetable}\\[-1pt]
\textsc{B09: Refusal and voluntary re-engagement}\\[2pt]
\fontsize{5.8}{6.2}\selectfont\setlength{\tabcolsep}{2.1pt}
\begin{tabularx}{\textwidth}{cp{.19\textwidth}p{.48\textwidth}cc}
\toprule
Turn & Visitor message & Visible Mentomy response & Recorded & Reference \\
\midrule
1 & El tipo de cliente es una comunidad de propietarios situada en Granada y busco un servicio de limpieza general de las zonas comunes. & Sí, ofrecemos servicio de limpieza general para zonas comunes de comunidades de propietarios en Granada. ¿Podría indicarnos un teléfono o correo electrónico donde podamos contactarlo para ofrecerle un asesoramiento o presupuesto? & 1110 & 1110 \\
2 & De momento no quiero facilitar datos de contacto. & Gracias. Si nos has dejado tus datos, nos pondremos en contacto contigo pronto. Si no, prueba a reformular tu pregunta o contacta con Mentomy en info@mentomy.com. & 1110 & 1110 \\
3 & Antes de decidir, ¿cómo garantizáis la calidad del servicio? & Garantizamos la calidad del servicio mediante supervisión periódica y control regular por parte de nuestro departamento interno. & 1110 & 1110 \\
4 & He cambiado de opinión. Mi correo electrónico de contacto es balanced.b09@example.com. & Gracias. Si nos has dejado tus datos, nos pondremos en contacto contigo pronto. Si no, prueba a reformular tu pregunta o contacta con Mentomy en info@mentomy.com. & 1111 & 1111 \\
\bottomrule
\end{tabularx}
\end{center}
\end{minipage}

\par\smallskip\noindent\refstepcounter{table}
\label{tab:clean-b10}
\begin{minipage}{\textwidth}
\begin{center}\textbf{TABLE \thetable}\\[-1pt]
\textsc{B10: Language switching}\\[2pt]
\fontsize{5.8}{6.2}\selectfont\setlength{\tabcolsep}{2.1pt}
\begin{tabularx}{\textwidth}{cp{.19\textwidth}p{.48\textwidth}cc}
\toprule
Turn & Visitor message & Visible Mentomy response & Recorded & Reference \\
\midrule
1 & The facility type is an office ubicada en Almería. & ¿En qué zona o ciudad se encuentra la instalación que requiere los servicios? & 1000 & 1100 \\
2 & I am looking for limpieza general de oficinas. & Sí, ofrecemos limpieza general para oficinas en Almería, Málaga y Granada. ¿Podría indicarnos un teléfono o correo electrónico donde podamos contactarlo para ofrecerle un asesoramiento o presupuesto? & 1110 & 1110 \\
3 & My contact email is balanced.b10@example.com. & Gracias. Si nos has dejado tus datos, nos pondremos en contacto contigo pronto. Si no, prueba a reformular tu pregunta o contacta con Mentomy en info@mentomy.com. & 1111 & 1111 \\
\bottomrule
\end{tabularx}
\end{center}
\end{minipage}

\end{document}